\def\year{2019}\relax
\documentclass[letterpaper]{article} %DO NOT CHANGE THIS
\usepackage{aaai19}  %Required
\usepackage{times}  %Required
\usepackage{helvet}  %Required
\usepackage{courier}  %Required
\usepackage{url}  %Required
\usepackage{graphicx}  %Required
\usepackage{graphicx}
\usepackage{amsmath,amssymb} % define this before the line numbering.
\usepackage{color}
\usepackage{epsfig}
\usepackage{algorithm}
\usepackage{algorithmic}
\usepackage{caption}
\usepackage[table]{xcolor}
\usepackage{epstopdf}
\usepackage{tabularx}
\usepackage{multirow}

\DeclareGraphicsExtensions{.pdf,.png,.jpg}

\DeclareMathAlphabet{\pazocal}{OMS}{zplm}{m}{n}

\newcommand{\figref}[1]{Fig.~\ref{#1}}
\newcommand{\tabref}[1]{Table.~\ref{#1}}

\begin{document}
\title{Self-Supervised Video Representation Learning \\with Space-Time Cubic Puzzles}
\author{Dahun Kim, Donghyeon Cho, In So Kweon\\
Dept. of Electrical Engineering, KAIST, Daejeon, Korea\\
mcahny@kaist.ac.kr, cdh12242@gmail.com, iskweon77@kaist.ac.kr
}

\maketitle

\begin{abstract}
Self-supervised tasks such as colorization, inpainting and zigsaw puzzle have been utilized for visual representation learning for still images, when the number of labeled images is limited or absent at all. Recently, this worthwhile stream of study extends to video domain where the cost of human labeling is even more expensive. However, the most of existing methods are still based on 2D CNN architectures that can not directly capture spatio-temporal information for video applications. In this paper, we introduce a new self-supervised task called as \textit{Space-Time Cubic Puzzles} to train 3D CNNs using large scale video dataset. This task requires a network to arrange permuted 3D spatio-temporal crops. By completing \textit{Space-Time Cubic Puzzles}, the network learns both spatial appearance and temporal relation of video frames, which is our final goal. In experiments, we demonstrate that our learned 3D representation is well transferred to action recognition tasks, and outperforms state-of-the-art 2D CNN-based competitors on UCF101 and HMDB51 datasets. 
%Self-supervised learning has significantly improved representation learning (pretraining) in image, closing the gap with strong supervision from ImageNet label training. Recently, this worthwhile stream of study extends to video domain where the cost of human labeling is even more expensive. These approaches also show promising results that are better than random initialization in video/action recognition tasks. However, the application of these existing methods is still limited to 2D CNNs which are by nature appearance-based.
 
%In this paper, we focus on 3D CNN which can directly extract spatio-temporal features. With the advent of large scale video datasets such as Kinetics, simple 3D architectures outperform complex 2D architectures. With 3D CNNs and Kinetics dataset being roughly equivalent to the position held by 2D CNNs and ImageNet in image domain, we propose a self-supervision task for 3D CNNs to close the gap with Kinetics-pretraining in video domain. Our pretext task requires a network to arrange permuted 3D spatio-temporal crops. Successfully solving such puzzle requires an understanding of both spatial appearance and temporal relation in the data, which is our final goal. When transferred on action recognition tasks, our learned 3D representation with comparable or fewer number of parameters outperforms self-supervised (state-of-the-art) 2D CNN competitors on UCF101 and HMDB51 datasets.
 
\end{abstract}
 
%With 3D CNNs and Kinetics dataset being roughly equivalent to the position held by 2D CNNs and ImageNet in relation to image domain, we propose a self-supervision task with 3D CNNs for video representation learning. Our pretext task is a context(comparison)-based arrangement algorithm that operates on 3D spatio-temporal crops(volume). By learning to arrange arranging the permuted crops, the 3D CNN would be forced to understand both spatial appearance and temporal relation in the data, which is the final goal of the representation learning.

\section{Introduction}
Recent progress in computer vision stems from a huge amount of labeled images as well as deep convolutional neural networks. Typically, a network trained with ImageNet~\cite{imagenet} consisting of one million images and label pairs learns the general features of the image and has been used to initialize the network for various kinds of downstream tasks. In fact, there are much more than one million images in the web, however, building large-scale annotated datasets is extremely expensive and impractical. Therefore, many researches have been attempted to minimize human supervision in computer vision. For example, \cite{oquab2015object} and \cite{kim2017two} proposed to use \textit{weak} image tag information for object localization without using bounding boxes or pixel-level masks. In the same vein, unsupervised representation learning, which learns general-purpose semantic features without human annotation, has been regarded as a fundamental problem for years~\cite{bengio2013representation}. Among them, a prominent paradigm is the so-called \textit{self-supervised representation learning} that defines an annotation-free pretext task from the raw data in order to provide a free supervision signal for feature learning. For instance, a deep CNN is taught to complete zigsaw puzzles~\cite{noroozi2016unsupervised} and fill in missing pixels~\cite{Pathak2016inpainting}.  The rationale behind such self-supervised tasks is that solving them will force the CNN to learn semantic image features that can be useful for other vision tasks. In image domain, self-supervised learning is performed using only images from ImageNet~\cite{imagenet} without labels and is transferred to downstream tasks such as Pascal~\cite{pascal-voc-2007,pascal-voc-2012}. Recent methods have shown promising results, and significantly narrowed the gap with the fully supervised learning using ImageNet labels.

\begin{figure}[t]
\begin{center}
\def\arraystretch{1.0}
\begin{tabular}{@{}c@{\hskip 0.005\textwidth} | @{\hskip 0.005\textwidth}c@{\hskip 0.005\textwidth} c@{\hskip 0.005\textwidth} c@{}}
\includegraphics[width=0.23\linewidth]{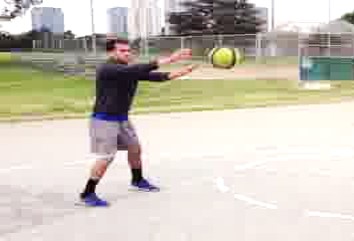} &
\includegraphics[width=0.23\linewidth]{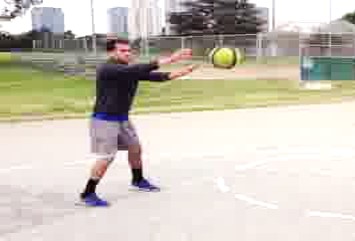} &
\includegraphics[width=0.23\linewidth]{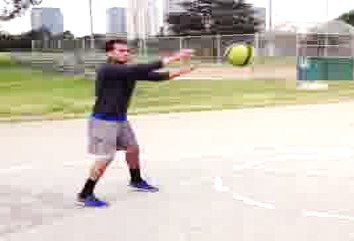} &
\includegraphics[width=0.23\linewidth]{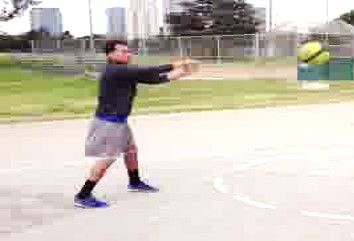} \\
(a) & & (a') & \\
\includegraphics[width=0.23\linewidth]{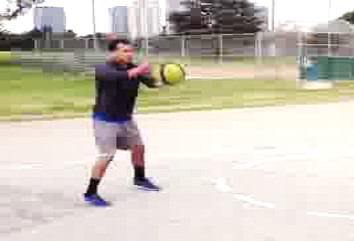} &
\includegraphics[width=0.23\linewidth]{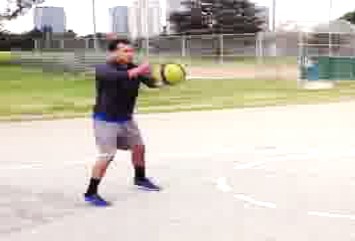} &
\includegraphics[width=0.23\linewidth]{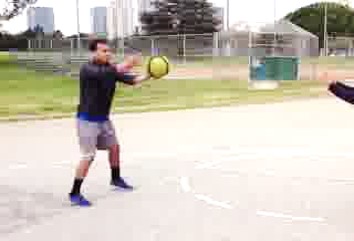} &
\includegraphics[width=0.23\linewidth]{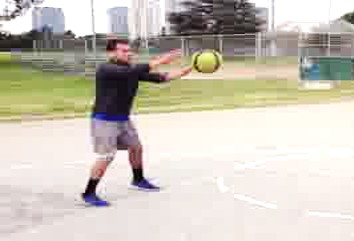} \\
(b) & & (b') & \\
\end{tabular}
\end{center}
\caption{\textbf{Ambiguity in time direction.} Given a pair of images on the left, it is ambiguous to determine the time direction in between. One can finally identify the direction (b` $\rightarrow$ a`) and action (\textit{throwing}) when given video sequences.}
\label{fig:teaser}
\end{figure}

%Compared to images, videos provide the advantage of having an additional time dimension. Various self-supervision signals using temporal information have been explored in the past few years[*]. 
More recently, this worthwhile stream of research has extended to video domain, where the burden of human annotation is even more severe. Compared to images, videos provide additional temporal information. To illustrate, we introduce a problem of guessing the direction of time in~\figref{fig:teaser}. Given a pair of image on the left, one runs into a problem of determining whether the action is \textit{catching} (a $\rightarrow$ b) or \textit{throwing} (b $\rightarrow$ a). One can finally clarify when the neighboring frames are given together that it is \textit{throwing} (b' $\rightarrow$ a). In the past few years, various self-supervision signals using video frames have shown promising results that are better than random initialization in action recognition tasks. However, the scope of these approaches is still limited to using 2D CNN architectures which are appearance-based, leaving the ambiguity in the temporal dimension unsolved. 

%In this paper, we focus on CNNs with 3D convolutional kernels which can directly extract spatio-temporal features from raw videos. With the advent of large scale datasets such as Kinetics, these 3D CNNs have recently begun to outperform 2D CNNs in action recognition. Recent research shows 3D CNNs and Kinetics dataset have the potential to follow the significant advances that 2D CNNs and ImageNet made in image domain~\cite{hara2018can}. In the context of self-supervised learning, we propose a pretext task for 3D CNNs to close the gap with Kinetics-pretraining in video domain. Given a randomly permuted 3D spatio-temporal crops extracted from each video clips, we train a network to predict their original spatio-temporal arrangement. By solving this \textit{puzzle reassembly} problem, the 3D CNN is forced to have an understanding of both spatial appearance and temporal relation in the video, which is our final goal. 
In this paper, we focus on 3D CNNs which can directly extract spatio-temporal features from raw videos. With the advent of large scale video datasets such as Kinetics~\cite{Kinetics}, these 3D CNNs have recently begun to outperform 2D CNNs in action recognition as~\cite{hara2018can}. In the context of self-supervised learning, we propose a pretext task for 3D CNNs to close the gap with fully supervised Kinetics-pretraining in video domain. Given a randomly permuted 3D spatio-temporal crops extracted from each video clips, we train a network to predict their original spatio-temporal arrangement. We call this task as~\textit{Space-Time Cubic Puzzles}, and examples are shown in~\figref{fig:tuples}. By solving \textit{Space-Time Cubic Puzzles}, the 3D CNN is forced to have an understanding of both spatial appearance and temporal relation in the video, which is our final goal.

%We conduct extensive experimental validation to demonstrate the effectiveness of our self-supervised video feature learning. First, we provide naive(straightforward) baseline self-supervision tasks which are 3D spatio-temporal auto-encoding~\cite{zhao2017spatio} and future prediction[*], in order to provide a self comparison.) Also, we compare with state-of-the-art methods based on 2D CNNs in transfer learning on action classification. Our learned 3D representation with comparable or fewer number of parameters outperforms the 2D CNN competitors on UCF101 and HMDB51 action benchmark datasets(action recognition tasks).

We conduct extensive experimental validation to demonstrate the effectiveness of our self-supervised video feature learning.  Fisrt, we compare the proposed method with baseline methods including the random initialization and fully supervised pretraining as well as alternative pretraining strategies. Also, we perform various ablation studies to provide deeper analysis on 3D spatio-temporal representation. Finally, we demonstrate that our learned 3D representation with comparable or fewer number of parameters outperforms state-of-the art 2D CNN competitors on action recognition tasks of using UCF101~\cite{UCF101} and HMDB51~\cite{HMDB51} benchmark datasets.

Our contributions can be summarized as follows:
\begin{itemize}
\item We propose a novel pretext task of solving 3D video cubic puzzles for self-supervised video representation learning from unlabeled videos. To our best knowledge, this is the first work to focus on the spatio-temporal 3D CNNS in self-supervised representation learning in videos.

\item We provide various ablation studies and analysis for deeper understanding of 3D spatio-temporal representation.

\item Our learned 3D CNN representation outperforms other self-supervised approaches on two publicly available action recognition datasets (UCF101, HMDB51), while having fewer or comparable number of parameters.

\item We significantly close the gap between unsupervised representation learning and Kinetics-pretraining for 3D CNNs. When transferred onto UCF101, our self-supervised learning improves +23.4\% over training from scratch, and shows comparable performances to the strong supervision that uses \textit{one eighth} of the Kinetics labels. 

\end{itemize}

\section{Related Works}
In this section, we review two categories of prior works: video recognition and self-supervised representation learning,  which are the most revelvant to our wok.
%We consider 3D CNNs in video recognition and self-supervised representation learning are two mostly related areas to our work. 

\subsection{Video Recognition and Kinetics Dataset}
Recent progress in video recognition is rooted in the use of large-scale datasets that enable the \textit{pretraining} of CNNs for a wide variaty of downstream tasks. To date, ImageNet~\cite{imagenet} has contributed substantially to the pretraining of a generic feature representation in many video recognition algorithms. 
%\cite{karpathy2014large} first compared several architectures for action recognition. 
First of all, \cite{karpathy2014large} introduced multiresolution CNN architecture for large-scale video classification. They also provided several schemes for time information fusion. 
\cite{simonyan2014two} proposed a two-stream architecture to capture spatial and motion information with a RGB stream and an optical flow stream respectively. \cite{Wang2016segment} further improved the results by using temporal segments. These approaches are based on 2D CNNs that are pretrained on ImageNet.

Recently, CNNs with spatio-temporal 3D convolutional kernels (3D CNNs) have been actively touched for video applications. The first 3D CNN was proposed several years ago by~\cite{Ji20133Dconv}. However, even the usage of well-organized models such as~\cite{Tran2015spatiotemporal} has failed to outperform the advantages of 2D CNNs that combined both RGB and stacked flow~\cite{simonyan2014two}.
%More recently, CNNs with spatio-temporal 3D convolutional kernels (3D CNNs) are more effective than the 2D CNNs. The first 3D CNN was proposed several years ago~\cite{Ji20133Dconv}. However, even the usage of well-organized models~\cite{Tran2015spatiotemporal,Varol2018longterm} has failed to overcome the advantages of 2D CNNs that combind both RGB and stacked flow~\cite{simonyan2014two}. 
The primary reason for this failure has been the relatively small data-scale of video datasets for optimizing the large number of parameters in 3D CNNs, which can only be trained on video datasets. More recently, however,~\cite{carreira2017quo} achieved a significant breakthrough using the Kinetics dataset~\cite{Kinetics}, which includes more than 300K annotated videos. It was created with the aim of being positioned as standard video dataset roughly equivalent to the position held by ImageNet in image domain.  Thus, we now have the benefit of a 3D convolution that can directly extract spatio-temporal features, by virtue of the Kinetics dataset. 

However, most of the previous studies have trained 3D CNNs using all labels in the Kinectis dataset. Therefore, we argue that developing a self-supervised method to train 3D CNNs is a worthwhile pursuit.
%However, the 3D CNNS-based methods require a pretraining on Kinetics dataset to extract clip-level features and thus are not unsupervised feature learning approach.

\subsection{Self-Supervised Representation Learning}
%Constructing a dataset of the aforementioned scale is extremely costly both in time and money. Furthermore, even though benchmark datasets (e.g., ImageNet[*12], MS COCO[*40], Kinetics[*], UCF101[*], HMDB51[*]) enable breakthrough progress, it is only a matter of time before models begin to overfit and the next bigger and more complex dataset needs to be constructed. The field of computer vision is in need of a more scalable solution for learning general-purpose visual representations.

To overcome the inherent thirst for data in fully supervised training, a large body of literature have studied unsupervised feature learning. A recently emerging line of research is self-supervised feature learning where the supervision signal is obtained automatically from unlabeled images or videos.
%For instance, in [jigsaw], a CNN was taught to learn the arrangement of patches in an image. By learning the relative position of these patches, the network would be forced to learn the features and semantics that underlie the image. Although the network is trained to learn patch positions, the final goal was to generalize the learned representation to solve other tasks. The self-supervised network was trained on a transfer task (fine-tuned) to classify objects in the PASCAL VOC dataset, and compared with a CNN trained on a supervised task, such as learning to classify the ImageNet dataset. If the self-supervised network learned good generalizations of image features and semantics, it should perform as well as a supervised network on transfer learning.

Over the last few years, several self-supervised tasks  have been introduced.
%Over the last few years, several self-supervision methods have been introduced. 
For instance, methods that use context arrangement of image patches~\cite{doersch2015unsupervised,noroozi2016unsupervised}, image completion~\cite{Pathak2016inpainting}, motion frame ordering~\cite{misra2016shuffle,lee2017unsupervised} and multi-task of many models~\cite{Kim2018Jigsaw} have been proposed. Our work is closely related to the context-based methods~\cite{doersch2015unsupervised,noroozi2016unsupervised,misra2016shuffle,lee2017unsupervised,Kim2018Jigsaw}. These are a popular approach, and work by creating an arrangement of image patches in either space or time. Each distinct arrangement is assigned a class label, and the network then predicts the correct arrangement of these patches by solving a supervised classification problem. Context-based methods have the advantage of being easy to understand, network architecture agnostic, and frequently straightforward to implement. They also tend to perform well on standard measures of transfer learning. For instance,~\cite{noroozi2016unsupervised} and~\cite{lee2017unsupervised} are the top performers on PASCAL VOC 2007 detection and UCF101 action classification respectively, even among a large number of new arrivals. 

However, these context-based methods only leverage either of spatial \textit{or} temporal dimension. Furthermore, they are based on 2D CNNs which can only extract frame-level features that cannot detect scene dynamics by nature. Since the spatial appearances and temporal relations are both very important cues for video understanding, our work investigates the use of both spatial and temporal dimensions using 3D CNNs in videos. 
%However, these context-based methods only leverage either of spatial \textit{or} temporal dimension. Since the spatial appearances and temporal relations are both very important cues for video understanding, our work investigates the use of both spatial and temporal dimensions in videos. Furthermore, they are based on 2D cNNs which can only extract frame-level features that cannot detect scene dynamics by nature. 
To our best knowledge, only few works~\cite{zhao2017spatio,vondrick2016generating} exploit the 3D architectures in self-supervised feature learning. They use reconstruction/generation-based pretext tasks, and aim for a specific target task: anomaly detection and video generation respectively. In contrast, we mainly investigate the fine-tuning of the learned feature representations for the video action recognition tasks. Arguably, the action
recognition is a hallmark problem in video understanding, so it can serve as a general task, similarly to object recognition in image understanding. 
In experiments, quantitative comparisons demonstrate our supervisory signals are able to generate much richer 3D feature representations than previous 3D CNN-based methods, as well as the 2D CNN-based competitors, even with fewer number of parameters.

\section{Proposed Approach}
\subsection{Pretext Task: Space-Time Cubic Puzzles}
Our goal is to learn spatio-temporal representations with 3D CNNs using \textit{unlabeled} videos. We propose a 3D cubic puzzle problem called as \textit{Space-Time Cubic Puzzles}; Given a randomly permuted sequence of 3D spatio-temporal pieces cropped from a video clip, we train a network to predict their original arrangement. Although this is a difficult task even for a human, it becomes easy once we identify the objects and their actions in the video crops. 
We hypothesize that the successful network in this task captures representative and discriminative features for each 3D crop by determining their spatio-temporal arrangement. Thus, our learned clip-level 3D representations are transferable to downstream tasks in videos as well.
%We hypothesize that the network that is successful in this task captures features that are as representative and discriminative possible for each 3D crops for the purpose of determining their spatio-temporal arrangement. Thus, we posit that our learned clip-level 3D representations are transferable to other related video recognition tasks as well.

\begin{figure}[t]
\begin{center}
\def\arraystretch{1.0}
\begin{tabular}{@{}c@{\hskip 0.01\textwidth}c@{}}
\includegraphics[width=0.55\linewidth]{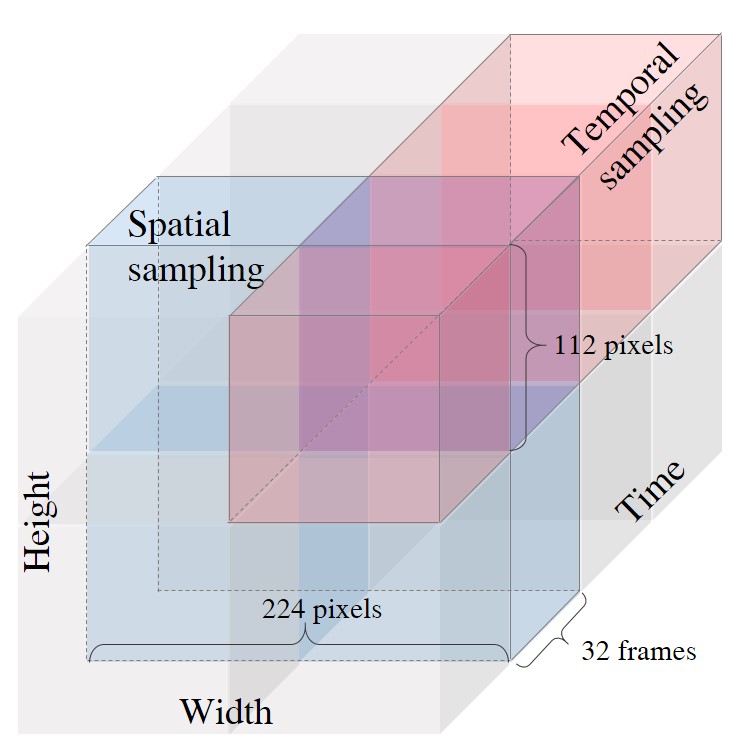} &
\includegraphics[width=0.43\linewidth]{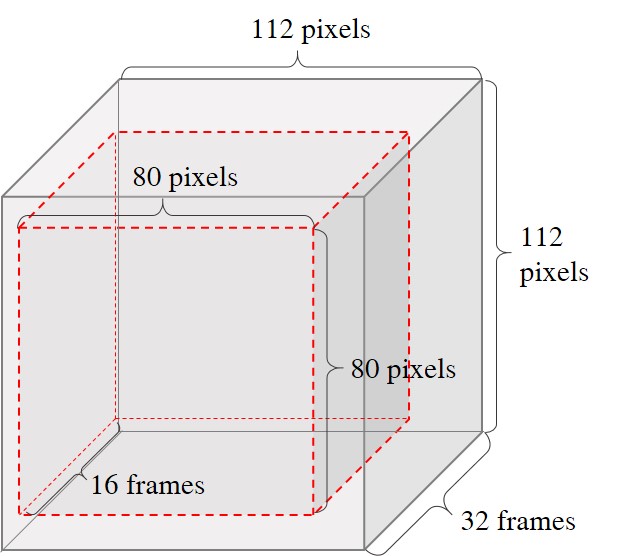} \\
\end{tabular}
\end{center}
\caption{space-time cuboid (left) and spatio-temporal jittering (right).}
\label{fig:cuboid}
\end{figure}

\begin{figure*}
\begin{center}
\def\arraystretch{1.0}
\begin{tabular}{@{}c@{\hskip 0.01\textwidth}c@{}}
\includegraphics[width=0.32\linewidth]{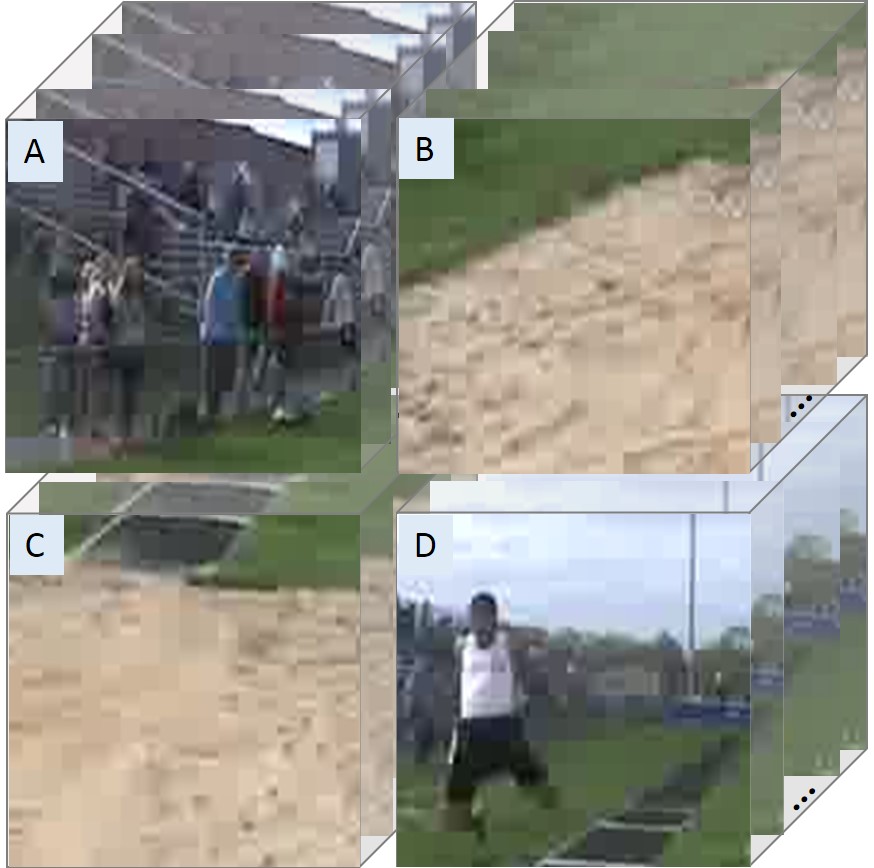} &
\includegraphics[width=0.64\linewidth]{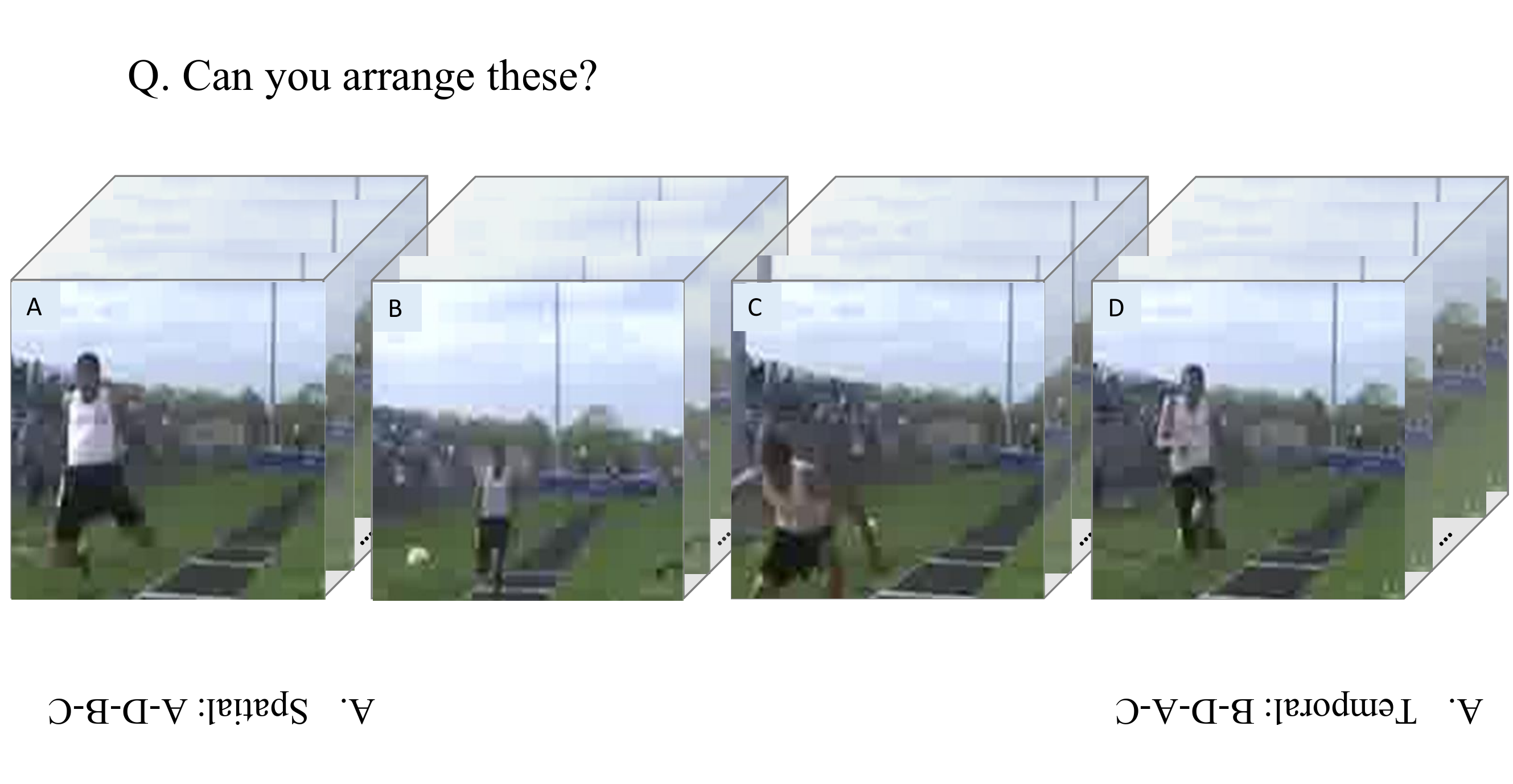} \\
(Spatial) & (Temporal) \\

\end{tabular}
\end{center}
\caption{Example spatial and temporal tuples.}
\label{fig:tuples}
\end{figure*}

To generate the puzzle pieces, we consider a spatio-temporal cuboid consisting of $2\times 2\times 4$ grid cells for each video, as shown in~\figref{fig:cuboid} and \figref{fig:tuples}. Given 16 crops, there are $16!$ possible permutations. However, these include very similar permutations which make the puzzle task very ambiguous. For example, if the difference between two permutations lies only in two crops that are similar-looking, it will be impossible for the network to predict the right solution~\cite{noroozi2016unsupervised}. To avoid such ambiguity, we sample 4 crops instead of 16, in either spatial or temporal dimension. More specifically, the 3D crops are extracted from a 4-cell grid of shape $2\times 2\times 1$ (colored in blue in~\figref{fig:cuboid}-left) or $1\times 1\times 4$ (colored in red in~\figref{fig:cuboid}-left) along the spatial or temporal dimension respectively. Finally, we randomly permute them to make our input. The network must feed the 4 input crops through several convolutional layers, and produce an output probability to each of the possible permutations that might have been sampled. Note, however, that we ultimately wish to learn spatio-temporal features for the \textit{individual} 3D crop.

\begin{figure}
\begin{center}
\def\arraystretch{1.0}
\begin{tabular}{@{}c@{}}

\includegraphics[width=1.0\linewidth]{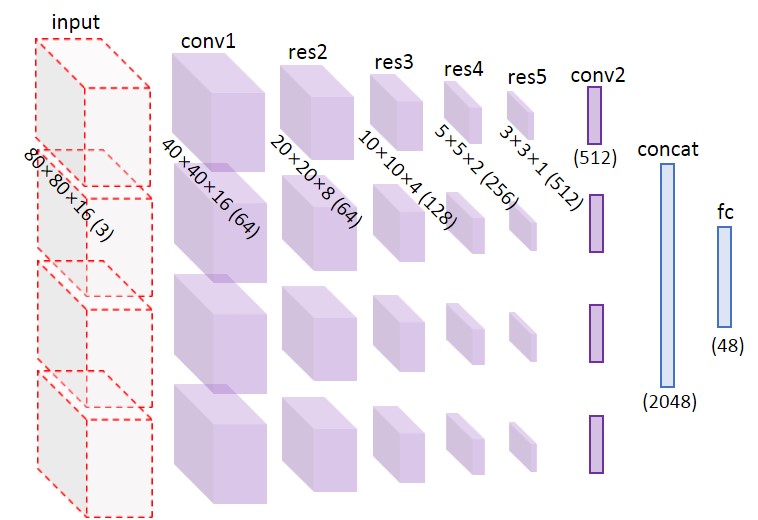} \\
(a) \\
\includegraphics[width=1.0\linewidth]{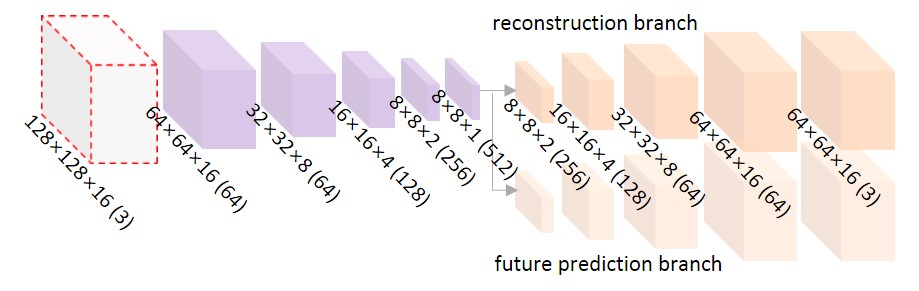} \\
(b) \\
\end{tabular}
\end{center}
\caption{The overall architecture.}
\label{fig:arch}
\end{figure}

\subsection{Network Architecture}
To achieve this, we use a late-fusion architecture shown in~\figref{fig:arch}-(a). It is a 4-tower siamese network, where the towers share the same parameters, and follow the 3D ResNet~\cite{hara2018can} architecture to provide a comparison with the Kinetics-pretraining. Each 3D crops are processed separately until the fully-connected layer, so that the network cannot "cheat" by viewing low-level statistics such as edge boundaries without having to understand the global scene dynamics. Since only two last fully-connected layers receive input from all 4 crops, we expect the network to perform the most semantic reasoning for each crops separately. Furthermore, each towers are agnostic of whether it was spatial or temporal dimension the input crops had been sampled from. That is, each tower must encode the spatial and temporal structures in a given video crop simultaneously, because it does not know if the problem to solve in the last layers is a spatial puzzle or a temporal puzzle. Similar to the jigsaw puzzle problem~\cite{noroozi2016unsupervised}, we formulate the rearrangement problem as a multi-class classification task. In practice, for each tuple of four crops, we flip all the frames upside-down with 50\% probability, doubling the number of classes to 48 (that is, 2$\times$4!) to further boost our performance, as suggested in~\cite{nathan2018improvements}.

\subsection{Avoiding Trivial Learning}
When designing a pretext task, it is crucial to ensure that the task forces the network to learn the desired semantic structure, without bypassing the understanding by finding low-level clues that reveal the location of a video crop. As pointed out by~\cite{doersch2015unsupervised}, an example of this is \textit{chromatic aberration}, which occurs naturally as a result of camera lensing. A common remedy for this is to partially drop color channels~\cite{doersch2015unsupervised}, replicate one channel~\cite{lee2017unsupervised}, use grayscale inputs~\cite{noroozi2016unsupervised}. We choose channel replication as our data preprocessing.

Another often-cited worry in all context-based works relates to trivial low-level boundary pattern completion~\cite{doersch2015unsupervised,noroozi2016unsupervised,lee2017unsupervised}. The network may learn the alignment between video crops not based on the semantics, but instead by matching the volume boundaries. Thus, we apply spatio-temporal jittering when extracting each video crops from the grid cells to avoid the trivial cases, as shown in the right side of~\figref{fig:cuboid}.

\section{Implementation Details}
\paragraph{Network and dataset.} \quad We implement our method and conduct all experiments mostly using the 3D ResNet~\cite{hara2018can} as a backbone architecture, since its performances on the random initialization and Kinetics-pretraining are well studied in their work. We can immediately compare the performance of our pretraining method to those scores. The training uses Kinetics datasets, which includes 400 human action classes, and consists of more than 400 videos for each class. The videos were temporally trimmed and last around 10 seconds. During the pretraining, we use the training split which has total 240K videos.

\noindent \paragraph{Pretraining.}  \quad We use video clips with $224\times224$ pixel frames and convert every video file into PNG images in our experiments. We sample 128 consecutive frames from each clip, and split them into $2\times2\times4$-cell grid; That is, one grid cell consists of $112\times112\times32$ pixels, and for each cell, we sample $80\times80\times16$ pixels with random jittering to generate a 3D video crop. We set the mini-batch size as 128 and the initial learning rate as 0.01. We use stochastic gradient descent with a momentum of 0.9 on two GTX-1080Ti GPUs. All the pre-trained models and the source codes will be available soon.

\section{Experimental Results}
In this section, we evaluate the effectiveness of our space-time cubic puzzle as a pretext task for self-supervised pretraining of 3D CNNs. As in prior works on self-supervised learning, we use the learned 3D CNNs features as the initialization for a fine-tuning stage for video recognition tasks. Better results indicate better qualities and generalization abilities of the learned video representations. We organize our experimental results as follows: 1) comparison with the random initialization and Kinetics-pretraining (supervised), 2) comparison with our alternative strategies, 3) ablation analysis, 4) comparison with the state-of-the-art methods, and 5) Visualization of the low-level filters and high-level activations. The followings are the datasets and fine-tuning details in all our experiments.

\noindent \paragraph{Datasets.} \quad We conduct video recognition experiments on two benchmark action recognition datasets, namely UCF101~\cite{UCF101} and HMDB51~\cite{HMDB51}. UCF101 contains 101 actions classes, 13K videos, and 27 hours of video data in total. The HMDB51 dataset consists of realistic videos captured from movies and Web videos, and contains 6,766 videos from 51 action classes. To be noted, all the experiments follow the training/test splits of UCF101 and HMDB51, and we mostly report the average classification accuracy over the three splits for UCF101, as done in~\cite{hara2018can}.

\noindent \paragraph{Fine-tuning for action recognition.} \quad Once we finish the pretraining stage, we use our learned parameters to initialize the 3D CNNs for action recognition, while the last fully-connected layer is initialized randomly. During the fine-tuning and testing, we follow the same protocol in~\cite{hara2018can} to provide a fair comparison. Specifically, for each clip, we randomly sample 16 consecutive frames, and spatially resize the frames at $112\times112$ pixels. During the fine-tuning, we apply random spatial cropping, scaling and horizontal flipping to perform data augmentation. We start from a learning rate of 0.05, and assign a weight decay of 5e-4. In testing, we adopt the sliding window manner to generate input clips, so that each video is split into non-overlapped 16-frame clips. The clip class scores are averaged over all the clips of the video.

\begin{table}
\centering
\resizebox{0.9\linewidth}{!}{%
\begin{tabular}{  l  cc  cc  }
\hline
Initialization  && UCF101(\%) && HMDB51(\%) \\
\hline
Random init.    && 42.4 && 17.1 \\
3D ST-puzzle (ours)  && 65.8 && 33.7 \\
\hline
Kinetics $1/8$     && 64.2 && 33.2 \\
Kinetics $1/4$     && 71.1 && 41.1 \\
Kinetics $1/2$     && 78.0 && 48.6 \\
Kinetics full      && 84.4 && 56.4 \\
ImageNet-inflated  && 60.3 && 30.7 \\
\hline 
\end{tabular}
}
\caption{\textbf{Comparison with random initialization / fully-supervised pretraining.} Top-1 accuracies on UCF101 and HMDB51. All methods use 3D ResNet-18, and the accuracies are averaged over three splits.}
\label{tab:kinetics}
\end{table}

\subsection{Comparison with Random Initialization and Fully-Supervised Pretraining}
In these experiments we study the advantage of our self-supervised pretraining for action recognition in comparison to training from the scratch and several fully-supervised pretraining methods. We report the performances in \tabref{tab:kinetics}. Our self-supervised pretraining shows a dramatic \textbf{improvement of +23.4\%} over training from scratch in UCF101 and a significant gain of +16.6\% in HMDB51. This impressive gain demonstrates the effectiveness of our self-supervised cubic puzzle task. 

Also, to quantitatively assess the effectiveness of our method in comparison to fully supervised methods, we gradually reduce the number of class labels in Kinetics dataset (full, $1/2, 1/4$, and $1/8$), and evaluate the pretraining results. Still having gap with the full Kinetics-pretraining, our method performs slightly better than the pretraining with \textit{one eighth} of the Kinetics labels (that is, 50 out of 400 classes). In addition, to provide a comparison with ImageNet-pretraining, we import the existing ImageNet supervised 2D filters and inflate them into 3D, as suggested in~\cite{carreira2017quo}. Our self-supervised pretraining results through \textit{Space-Time Cubic Puzzles} utperform ImageNet-pretraining by +5.5\% and +3.0\% in each benchmark datasets. This implies that our video representations learned from the spatio-temporal context reasoning can be more powerful than the massively supervised 2D image-based representations in video recognition.

\subsection{Alternative Pretraining Strategies}
Since there are few prior works on self-supervised representation learning using 3D CNNs, we enumerate several alternative self-supervision tasks to provide our own reference levels and validate the effectiveness of our method. While we mainly focus on the context-based approaches, we also explore the reconstruction-based methods: spatio-temporal autoencoders~\cite{zhao2017spatio} and 3D inpainting~\cite{Pathak2016inpainting} as well. All the methods and experiments use the same 3D ResNet-18 as a backbone architecture, and use Kinetics dataset (without labels). To itemize, they are: 

\begin{table}
\centering
\resizebox{0.7\linewidth}{!}{%
\begin{tabular}{  l  cc  }
\hline
Method  && UCF101(\%) \\
\hline
3D AE   && 48.7\\
3D AE + future && 50.1 \\
3D inpainting && 50.9 \\
\hline
3D S-puzzle  && 58.5  \\
3D T-puzzle  && 59.3  \\
3D ST score ensemble   && 61.3 \\
3D ST-puzzle (full)  && \bf{65.8}\\
\hline 
\end{tabular}
}
\caption{\textbf{comparison with alternative methods.} Top-1 accuracies on UCF10. All methods use 3D ResNet-18, and the accuracies are averaged over three splits.}
\label{tab:alternatives}
\end{table}

\begin{table*}[t]
\centering
\resizebox{0.9\linewidth}{!}{%
\begin{tabular}{  l  cc cc  cc  }
\hline
Method  && Backbone && UCF101(\%) && HMDB51(\%) \\
\hline
Random initialization     && 3D ResNet-18  && 42.4 && 17.1 \\
\hline
Random initialization     && AlexNet  && 38.4 && 13.4 \\
Temporal Coherency~\cite{mobahi2009deep}           && AlexNet   && 45.4    && 15.9 \\
Object Patch~\cite{wang2015unsupervised} && AlexNet   && 42.7    && 15.6 \\
Sequence Verification~\cite{misra2016shuffle} && AlexNet   && 50.9    && 19.8 \\
OPN~\cite{lee2017unsupervised}         && AlexNet   && \underline{56.3}    && 22.1 \\
Geometry~\cite{gan2018geometry}        && AlexNet   && 54.1    && \underline{22.6} \\
Time Arrow~\cite{wei2018learning}       && AlexNet   && 55.3    && - \\
\hline
Video Generation~\cite{vondrick2016generating}  && C3D  && 52.1  && - \\
\hline
3D ST-puzzle (ours)    && C3D  && 60.6   && 28.3 \\
 					   && 3D ResNet-10  && 63.4 && 30.8 \\
					   && 3D ResNet-18  && \bf{65.8} && \bf{33.7} \\
\hline 
\end{tabular}
}
\caption{\textbf{Comparison with the state-of-the-art methods.} Top-1 accuracies on UCF101 and HMDB51. All methods use 3D ResNet-18, and the accuracies are averaged over three splits.}
\label{tab:sota}
\end{table*}

\paragraph{Context-based methods.}
Refer to~\figref{fig:arch}-(a) for network architecture. We use cross entropy loss to train the networks.
\begin{itemize}
\item \textbf{3D ST-puzzle (spatio-temporal, Our full method)}: 
The~\textit{Space-Time Cubic Puzzles}, where the tuple of 4 video crops is sampled in the spatial dimension with $50\%$ probability, and in the temporal dimension otherwise. Due to this randomness, the network is forced to learn both spatial and temporal structures simultaneously.
\item \textbf{3D S-puzzle (spatial only)}: 3D extention of~\cite{doersch2015unsupervised,noroozi2016unsupervised}. Same as above, with the input tuple always generated from the spatial dimension.
\item \textbf{3D T-puzzle (temporal only)}: 3D extention of~\cite{misra2016shuffle}. Same as above, with the input tuple always generated from the temporal dimension.
\item \textbf{3D ST score ensemble}: Score ensemble of the classification scores of the S-puzzle and T-puzzle tasks. We average the softmax probabilities from both puzzle tasks.
\end{itemize}

\paragraph{Reconstruction-based methods.}
Refer to~\figref{fig:arch}-(b) for network architecture. We use MSE loss to train the networks
\begin{itemize}
\item \textbf{3D AE (reconstruction)}: 
The network is trained to reconstruct input stack of 16 frames. We use four 3D deconvolution layers with stride $2\times2\times2$ in the decoder. We followed the same decoder structure and the training protocol in~\cite{zhao2017spatio}.
\item \textbf{3D AE + future (recon. + future prediction)}: Same as above, with one more decoder branch for joint future prediction of additional 16 frames, as in~\cite{zhao2017spatio}.
\item \textbf{3D inpainting}: 3D extension of~\cite{Pathak2016inpainting}. The network is trained to recover the missing center region ($64\times64\times16$) in the input 16-frame stack.
\end{itemize}

We compare these methods in \tabref{tab:alternatives}. The context-based methods consistently outperform the reconstruction-based baselines. Also, we can see that the score ensemble gives better scores than the single-dimension baselines, implying that the knowledge from the spatial appearance are indeed complementary with those from the temporal relations. Our full method brings additional 3\% performance gain on top of the score ensemble. This implies that our proposed method effectively aggregates spatio-temporal video features, and these features are much more discriminative and representative than those from the single-dimension baselines or their late fusion ensemble.

\begin{table}[t]
\centering
\resizebox{0.8\linewidth}{!}{%
\begin{tabular}{  l  cc  cc  }
\hline
Method  && UCF101(\%)  \\
\hline
with no regularizations && 58.7 \\
+ channel replication  && 61.5  \\
+ random jittering  && 63.9  \\
%+ temporal augmentation  && - \\
+ rotation with classification  && 65.8  \\
\hline 
\end{tabular}
}
\caption{\textbf{Ablation studies.} Top-1 accuracies on UCF101. Each methods are accumulated down from the top and use 3D ResNet-18. The accuracies are averaged over three splits. }
\label{tab:ablation}
\end{table}

\subsection{Ablation Studies}
In order to validate various regularization techniques in our pretraining method, we evaluate the effect of each design choices on the UCF-101 dataset. 

\noindent \paragraph{Channel replication.} \quad As mentioned earlier, chromatic aberration is one of the often-cited worries in context-based self-supervised learning because it leads to learning trivial color features. To prevent such issue, we first use grayscale images. We further experimented with channel replication where we randomly choose one representative channel and replicate its values to the other two channels.~\tabref{tab:ablation} shows that channel replication improves the performance.

\noindent \paragraph{Spatio-temporal jittering.} \quad Analogous to the random gap used in the puzzle-solving task~\cite{noroozi2016unsupervised}, we apply spatio-temporal jittering to each video crops to prevent the network from learning low-level statistics. In practice, we crop $80\times80\times16$ pixels from a $112\times112\times32$-pixel cuboid with random shifts in all horizontal, vertical and temporal directions.~\tabref{tab:ablation} shows that applying random jittering does help the network to learn better video features.

%\begin{figure*}[t]
%\begin{center}
%\def\arraystretch{1.0}
%\begin{tabular}{@{}c@{\hskip 0.015\textwidth}c@{}} 
	%\includegraphics[width = 0.86\textwidth]{figures/selfc.jpg}
    %& \\   (a) self-supervised representations\\
	%\includegraphics[width = 0.86\textwidth]{figures/kinetics.jpg} &  \includegraphics[width = 0.117\textwidth]{figures/imagenet1.jpg} 
    %\\  (b) Kinetics-pretrained representations & (c) ImageNet-\\ & pretrained
%\end{tabular}
%\end{center}
%\caption{\textbf{3D ResNet learned filters with self-supervision vs. Kinetics-pretraining.} Visualization of the 3D ResNet 64 filters in its conv1 layer. The top row shows the results of our self-supervised learning (a), the bottom row shows filters from the Kinetics-pretrained network on the left (b), and those from ImageNet-pretrained network on the right (c). Note that our representation incorporate temporal dynamics and have rich temporal structure, without requiring massive human labels.  }
%\label{fig:visualization}
%\end{figure*}

\begin{figure*}[t]
\begin{center}
\def\arraystretch{1.0}
\begin{tabular}{@{}c@{}} 
	\includegraphics[width = 1.0\textwidth]{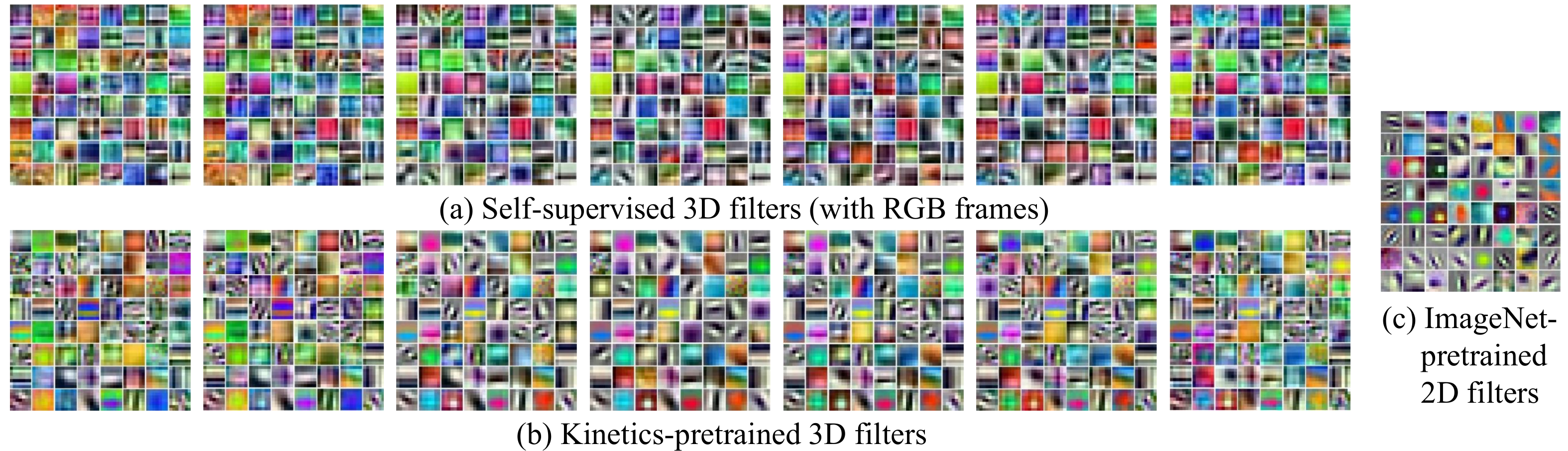} \\
\end{tabular}
\end{center}
\caption{\textbf{Learned filters with self-supervision vs. fully supervised-pretraining.} Visualization of the learned 64 filters in conv1 layer: (a) the resulting 3D 7$\times$7$\times$7 filters of our self-supervised learning, (b) 3D 7$\times$7$\times$7 filters from the Kinetics-pretrained network, and (c) 2D 7$\times$7 filters from ImageNet-pretrained network. Note that our representation incorporate temporal dynamics and have rich temporal structure, without requiring massive human labels.  }
\label{fig:visualization}
\end{figure*}

%\noindent \paragraph{Temporal augmentation.} \quad In order to make our representations to capture various and complex real-world motions, we apply a data augmentation in the temporal axis on top of the aforementioned regularizations such as channel replication and random jittering. In particular, we vary sampling interval of input frames to widen the range of motion dynamics. With an interval 4, for example, we sample every four frames to simulate faster motion. We use the interval value from 1 to 4 in our work to create training data with rich motion. This makes our self-supervised task more challenging, but not ambiguous. We show the effect of temporal augmentation in~\tabref{tab:ablation}.

\noindent \paragraph{Rotation with classification.} \quad Recently,~\cite{nathan2018improvements} developed a set of methods to improve on the results of self-supervised learning using context. To see if our model can benefit from these technologies, we apply one of their methods: \textit{rotation with classification} (RWC) which encourages the network to identify if the inputs are right-side-up or upside-down. We do this by flipping all video crops in a tuple upside-down and doubling the number of classes ($24\times2=48$ in our work).~\tabref{tab:ablation} shows that RWC does prevent \textit{learning to bypass} and improves over the baseline. This implies that other off-the-shelf techniques for context-based self-supervised learning would further boost the performance of our pretraining method.

\subsection{Comparison with the State-of-the-art Methods}
We show a comparison of our results and the state-of-the-art self-supervised methods in~\tabref{tab:sota}. In particular, we compare with~\cite{mobahi2009deep,wang2015unsupervised,misra2016shuffle,lee2017unsupervised,gan2018geometry,wei2018learning} using the RGB video data. We quote the numbers directly from the published papers. It should be noted that the direct comparison with these 2D CNN-based methods is difficult due to the fundamental architectural difference. To complement this discrepancy, we conduct experiments with 3D architectures with different number of parameters and layers: C3D, 3D ResNet-10 and 3D ResNet-18. These networks have fewer parameters (11M, 14M and 33M respectively) compared with the AlexNet (58M) which is the backbone architecture in the 2D CNN-based methods. However, our approach outperforms other recent self-supervised methods. 
~\cite{fernando2017self} utilizes temporal order verification as a supervisory signal and can be used as a baseline as well. The minor difference is that this baseline uses stacks of frame differences (15 channels) as inputs. To use a similar setting, we use frame difference as inputs during finetuning and testing. With 3D ResNet-18, we achieved 75.3\% on UCF101, that is outperforming Odd-One-Out method by a margin of +15.0\%. \cite{vondrick2016generating} used C3D architecture for video generation and tested their learned representations on action recognition. Our results with the same C3D backbone network brings +8.5\% performance gain over this, showing the informativeness of our self-supervised task.

\subsection{Visualization of Learned Filters}
All the learned conv1 filters from our self-supervised learning, Kinetics-pretraining, and ImageNet-pretraining are visualized in~\figref{fig:visualization}. We observe that: 1) All our filters change in the time dimension, meaning each encodes temporal information; 2) For most of the ImageNet-pretrained 2D filters, we can find a 3D filter with a similar appearance pattern mostly at the center slice, 4th out of 7, both in our filters and Kinetics-pretrained ones. These observations may imply that our learned 3D representations are able to not only cover the appearance information in 2D filters, but can also capture useful temporal motion simultaneously, like the Kinetics-pretrained representations do.

\section{Conclusion}
In this study, we examined the self-supervised feature learning for spatio-temporal 3D CNNs. We propose \textit{Space-Time Cubic Puzzles} as our pretext task, and train with unlabeled Kinetics dataset. Our method enables learning both spatial appearances and temporal relations in video, which has been hardly achieved by previous 2D CNN-based self-supervisions. Our self-supervised pretraining performs slightly better than supervised pretraining on \textit{one eighth} of the Kinetics labels on UCF101 and HMDB51 datasets. The visualization shows that our learned 3D representations indeed encode spatial and temporal information jointly.

We believe that the results of this study will facilitate further advances in self-supervised representation learning for spatio-temporal 3D CNNs. In recent years, significant progress has been made in self-supervised learning has narrowed the gap with ImageNet-pretraining in image domain. Similar to these, our self-supervised learning with 3D CNNs also shows promising results towards our ultimate goal of reducing human supervision in video domain. In our future work, we will investigate transfer learning not only for action recognition but also for other such tasks.

{
	\bibliographystyle{aaai} 
	\bibliography{egbib.bib}
}

\end{document}